\documentclass[journal]{IEEEtran}

\ifCLASSINFOpdf
\else
\fi

\usepackage[pagebackref=true,breaklinks=true,letterpaper=true,colorlinks,bookmarks=false]{hyperref}

\usepackage[pdftex]{graphicx}
\usepackage{bm}
\usepackage{amsmath}
\usepackage{amssymb}
\usepackage{algorithm}
\usepackage{algorithmic}
\usepackage{mathrsfs}
\usepackage{enumitem} 
\usepackage{multirow}
\usepackage{tabularx}

\graphicspath{image}

\def\eg{\emph{e.g.}}
\def\etal{\emph{et al.}}
\def\ie{\emph{i.e.}}
\def\dD{$\mathbf{d_\Delta}$}

\begin{document}
\title{Revisiting EmbodiedQA: \\ A Simple Baseline and Beyond}

\author{Yu Wu, Lu Jiang, and Yi Yang
\thanks{(Corresponding author: Yi Yang.)

Y. Wu, and Y. Yang are with the Center for Artificial Intelligence, University of Technology Sydney, Ultimo 2007, NSW, Australia. (E-mail: yu.wu-3@student.uts.edu.au; yi.yang@uts.edu.au).

L. Jiang is with Google Research, Mountain View, California, USA. (E-mail: lujiang@google.com)}
}

\markboth{IEEE TRANSACTIONS ON IMAGE PROCESSING, 2020}%
{Shell \MakeLowercase{\textit{et al.}}: Bare Demo of IEEEtran.cls for IEEE Journals}

\maketitle

\begin{abstract}
In Embodied Question Answering (EmbodiedQA), an agent interacts with an environment to gather necessary information for answering user questions. Existing works have laid a solid foundation towards solving this interesting problem. But the current performance, especially in navigation, suggests that EmbodiedQA might be too challenging for the contemporary approaches. In this paper, we empirically study this problem and introduce 1) a simple yet effective baseline that achieves promising performance; 2) an easier and practical setting for EmbodiedQA where an agent has a chance to adapt the trained model to a new environment before it actually answers users questions. In this new setting, we randomly place a few objects in new environments, and upgrade the agent policy by a distillation network to retain the generalization ability from the trained model. On the EmbodiedQA v1 benchmark, under the standard setting, our simple baseline achieves very competitive results to the-state-of-the-art; in the new setting, we found the introduced small change in settings yields a notable gain in navigation.
\end{abstract}


\IEEEpeerreviewmaketitle

\section{Introduction}
\begin{figure}[t]
    \centering
    \includegraphics[width=\linewidth]{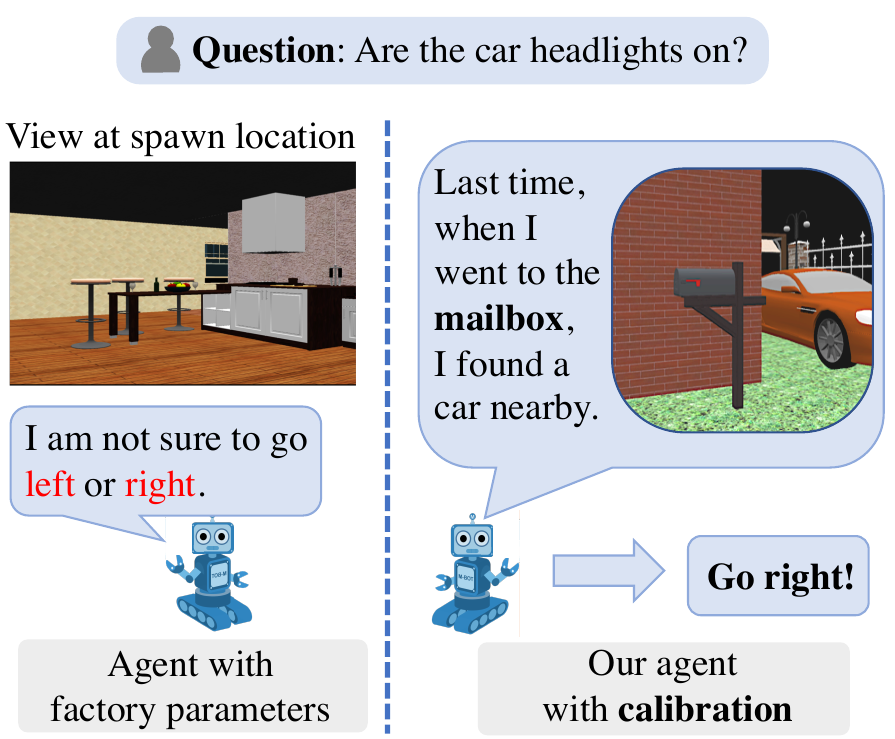}
    \caption{An illustration of the EmbodiedQA task under the standard evaluation setting (Left) and the calibration setting (Right). In the EmbodiedQA task, the agent should navigate to explore the environment and collect visual evidence to answer the asked question.
    In the figure, the asked question is shown on the top.
    To answer this question, the agent is given only the first-person view 
    and is expected to move closer to the car in an unseen environment.
    Under the standard setting, it is very difficult for an agent to navigate as it has never seen the environment before.
    In the new calibration setting, it is easier for the agent to choose the right direction as its model has been adapted to the new environment via the calibration.
    }
    \label{fig:introduction}
\end{figure}

 \IEEEPARstart{A}{} long-standing goal of artificial intelligence is to develop agents that can perceive and interact with the environment and communicate with humans in natural language. A representative research area is studying a goal-driven agent that can communicate with humans (language), perceive the environment (vision), and explore the space (taking actions). This paper focuses on a kind of such problem called Embodied Question Answering (EmbodiedQA)~\cite{embodiedqa}, a sub-field derived from Visual Question Answering (VQA), where users could ask an agent questions, and to answer these questions, the agent needs to perform actions to navigate the environment and collect evidence. A key difference to related problems, such as visual navigation~\cite{leonard2012directed,thrun2005probabilistic,mei2016navigational}, is that the agent is only given the first-person view and has no access to the global map of the environment nor the room/object layout in the environment. The example in Fig.~\ref{fig:introduction} illustrates this challenging setting where the agent needs to answer questions about an object at a random location in the environment.

The pilot works on EmbodiedQA in the literature~\cite{embodiedqa,yu2019multieqa,eqa_modular} have laid a solid foundation towards solving this interesting problem. These works have achieved promising results in terms of the final Question Answering (QA) accuracy. Despite the promising progress, EmbodiedQA remains to be very challenging to solve, and perhaps too challenging for the current approaches. A recent paper by Anand \etal~\cite{anand2018blindfold} reported that a blind agent, which is designed as a simple CNN Bag-of-words model to completely ignore the environment and visual information, can achieve the state-of-the-art QA performance. 
Besides fixing the statistical irregularities in the EmbodiedQA dataset as suggested in~\cite{anand2018blindfold}, on the other hand, Anand \etal's interesting finding inspires us to take a step back and \emph{empirically revisit the EmbodiedQA problem}.

First, Anand \etal's results suggest using the QA accuracy as the only evaluation metric is not sufficient. The evaluation metric on grounding seems, perhaps, equally important as the final QA accuracy. We can use the navigation distance in EmbodiedQA as the  grounding metric, which measures the change (in meters) during the navigation towards the target object. This idea agrees with contemporary works in VQA, where a system is designed to answer a question based on correct ``grounding snippets'' either spatially (\eg~in~\cite{zhu2017visual}) or temporally (\eg~in~\cite{liang2018focal}). Under the new criteria, a blind agent in~\cite{anand2018blindfold}, though enjoys a decent QA accuracy, still suffers from the random navigation accuracy.

Second, existing EmbodiedQA systems~\cite{embodiedqa,yu2019multieqa,eqa_modular} can be generally divided into a navigation module and a QA module. These two modules are first individually optimized and then jointly trained by Reinforcement Learning (RL). See Section~\ref{sec:related_works}. This is different from the simple baseline in~\cite{anand2018blindfold} as well as many other VQA models, where the model is often trained end-to-end by Stochastic Gradient Descent (SGD). This observation inspires us to seek the answer to the following question \emph{to what extend a true simple EmbodiedQA baseline is able to achieve in comparison to the state-of-the-art?} To this end, we introduce a new simple yet effective EmbodiedQA baseline. Different from existing RL-based approaches, it employs a simple policy model for navigation and a standard LSTM model for question answering. By virtue of this simple design, our network can be jointly trained by SGD in which both two modules can be optimized efficiently. Empirically, we found this simple baseline achieves competitive results on the EmbodiedQA benchmark both in terms of the final QA and the navigation accuracy. A notable benefit of our simple baseline is that it is simple to integrate with a wide variety of VQA models, most of which are trained by SGD as opposed to RL-based approaches.

Finally, our empirical results indicate that the QA bottleneck stems from the worse navigation in the unseen environment, \ie~an agent is hardly able to reach a point to observe the target object in a new environment, let alone answer the question. It suggests that the EmbodiedQA setting may be too challenging for the contemporary approaches.
Recall our goal is to understand the capability of simple EmbodiedQA methods. As a result, as opposed to developing more advanced navigation models, this paper introduces a slightly easier setting in which users are allowed to ask the agent a few rhetorical questions whenever it enters into a new environment. No extra supervision is needed since users already know and simply do not care about the answers to these questions. Our real goal is to warm up the agent in the new environment. It is a practical setting as the adaption can be conveniently finished via the same QA interface without changing the agent's pipeline.

To this end, we randomly place some reserved objects, called ``markers'', in a new environment and ask the agent questions about the colors of these markers in order to adapt the already trained model to the new environment. We call it \emph{calibration setting} as users can adapt the agent to a new environment in a similar way as calibrating a new camera. We found that this small change in setting considerably reduces the learning difficulty and yields notable gains in navigation. This setting shares a high-level similarity with visual navigation. But note in EmbodiedQA, we are not allowed to access to the environment structures (\ie, rooms, objects layout and global sketch map). 

We empirically compare our method to the state-of-the-art approaches on the EmbodiedQA v1 benchmark. 
Under the standard setting, we show that our simple baseline achieves very competitive results in terms of both QA and navigation accuracy. To the best of our knowledge, this is one of the first EmbodiedQA methods that is jointly optimized by SGD instead of Reinforcement Learning. In the new ``calibration'' setting, the results validate this small change in setting leads to a significant gain in navigation when the agent is further (30 and 50 action steps) away from the target. In summary, this paper presents an empirical study on the recent topic of EmbodiedQA. Our main observations can be summarized as:

\begin{itemize}
\item We propose a simple baseline method that can be jointly optimized \textit{without} Reinforcement Learning, which is competitive to the state-of-the-art EmbodiedQA methods. 

\item We introduce an easier and practical setting for EmbodiedQA.
We found this small change in setting yields a notable gain in navigation especially when the agent is further away from the target.

\item The marker generation and model adaptation may greatly influence the agent's performance. We found randomly placing a few marks and distill the knowledge from the trained model performs reasonably well in practice.

\end{itemize}

\section{Related work} \label{sec:related_works}

In the EmbodiedQA task, vision, language, and navigation action are combined together in building an intelligent goal-driven agent.
It origins from two widely studied tasks, the visual navigation task and the visual question answering task.
We first review related works for these two tasks and then extend the discussion to the EmbodiedQA task.

\subsection{Visual Navigation}
The problem of navigating in an environment based on visual input has long been studied in computer vision and robotics~\cite{leonard2012directed}.
Classical approaches~\cite{thrun2005probabilistic} constructed a 3D model of the environment based on visual observations, and then planed and selected paths based on this map.
Some early approaches~\cite{borenstein1989real,borenstein1991vector} required a prior global map or building an environment map on-the-fly.

Recently, deep reinforcement learning was developed and successfully applied to the visual navigation task.
Some works~\cite{jaderberg2016reinforcement,mirowski2016learning} utilized auxiliary tasks during training to improve navigation performance.
Others either took the recurrent neural network (RNN) to represent the memory~\cite{gupta2017cognitive,khan2018memory,mei2016navigational,parisotto2018} or predicted navigational actions directly from visual observations~\cite{Brahmbhatt_2017_CVPR,oh2017zero,zhu2017visual}.

In the visual language navigation task, the language inputs are instructions to teach the agent how to move.
However, in EmbodiedQA, the language input is a question about a target object, without any instructions on how to reach it. The visual navigation is only an intermediate step towards the end goal of question answering. Another difference is that the EmbodiedQA agent is only given the first-person view and has no access to the global map of the environment nor the room/object layout in the environment.
We study this problem in the context of EmbodiedQA.

\subsection{Visual Question Answering}
Recently, multimedia analysis~\cite{yang2012feature,yan2016image} has attracted a lot of research attention. 
Among them, visual question answering (VQA) is an interesting AI task that utilizes the input of both vision and language.
Many VQA approaches~\cite{agrawal2017vqa,malinowski2015ask,Wuqi2017pami,kazemi2017show,zhu2017uncovering} trained the model from the encoded image features and text features via the end-to-end network optimized by SGD.
A key research question in VQA is studying an effective multimodal module to combine the image and text for answer inference. 
Recent studies show the attention-based model~\cite{anderson2018bottom,fukui2016multimodal,xue2018tip,deng2018triplet,lu2016hierarchical,xu2016ask} is a promising direction for this purpose.
Some recent works~\cite{agrawal2018don, goyal2017making,liang2018focal,zhang2019interpretable} introduce visual grounding to the VQA models to avoid the language priors in training data. It is reasonable to expect VQA models that are answering questions for the ``right reasons''.

Existing VQA problems mainly focus on the static vision language dataset~\cite{agrawal2017vqa,gao2015you,goyal2017making,krishna2017visual,zhu2016visual7w}, where the agent answers questions based on a passive vision input, \ie, without the ability to interact with the environment.
In comparison, EmbodiedQA allows an agent to explore the space (take actions) before answering questions. It is worth emphasizing that since this paper is studying the capability of a simple method for EmbodiedQA, we use the standard VQA model for answer inference. The proposed model is compatible with advanced VQA models, and it is simple to incorporate existing VQA models.

\subsection{Embodied Question Answering}
Embodied Question Answering (EmbodiedQA) is a QA task that requires an agent to interact with a dynamic visual environment. 
This task is also relevant to vision-language grounding~\cite{Huang_2018_CVPR,mao2016generation}.
A common framework is to divide the task into two sub-tasks: an intermediate navigation task and a downstream QA task.
The navigation module is essential since it is a prerequisite to seeing an object before answering questions about it.
Previous works~\cite{embodiedqa,eqa_modular} focused on the navigation and used the same QA model for comparison. 

For example, Das~\etal~\cite{embodiedqa} proposed Planner-Controller Navigation Module (PACMAN), which is a hierarchical structure for the navigation module, with a planner to select actions (directions) and a controller that decide how far to move following this action.
The navigation module and visual question answering model are first trained individually and then jointly trained by REINFORCE~\cite{williams1992simple}.
Similarly, Gordon~\etal~\cite{gordon2018iqa} proposed the Hierarchical Interactive Memory Network, consisting of a hierarchy of controllers that operate at multiple levels of temporal abstraction and a rich semantic memory that aids in navigation, interaction, and question answering.
Das~\etal~\cite{eqa_modular} further improved the PACMAN model by Neural Modular Control (NMC) where the higher-level master policy proposes semantic subgoals to be executed by sub-policies.
Different from their method, we design a training pipeline to jointly optimizes navigation and QA modules \textit{without} Reinforcement Learning, which seems to be more effective.

EmbodiedQA may be too challenging to solve by current approaches. Previous works~\cite{embodiedqa,eqa_modular} achieve poor navigation performance, especially for the medium and long distance questions. This is because the agent is hardly able to see the target object in a new environment before it starts answering the question. Fortunately, a recent study from Yu~\etal~\cite{yu2019multieqa} proposed a  multi-target EmbodiedQA task to reduce this difficulty. In this setting, questions have multiple targets, and users have a chance to provide guidance to the agent for conducting the EmbodiedQA task.
Likewise, we tackle this issue but from a different angle. We study a slightly different setting that turns out greatly reducing the difficulty for navigation learning.
We propose a proxy task to help the agent adapt the learned model to the new environment using a few questions about the markers.
Building a global map for the unseen environment during the exploration is also a feasible solution but is out of the scope of the EmbodiedQA task.

\section{Background}
In the EmbodiedQA task, the agent is spawned at a random location in an environment. 
The agent has a single egocentric camera mounted at a fixed height, observing the environment by RGB vision input in the first person view. Note that it is not allowed to access global information (map, location coordinates) or the structure of the environment (rooms and objects).
Given an input question $q$ about a target object $T$, the agent needs to navigate in the environment to find the necessary information of the target to answer the question.
The agent can take action from the navigation actions space $\mathcal{A}$.
In the EQA v1 dataset, $\mathcal{A}$ consists of four actions, \ie, ``move forward'', ``turn left'', ``turn right'', and ``stop navigation and start to answer the question''.
The moving distance and rotation angle for one action step is predefined as 0.5 meters and 9 degrees, respectively.

Generally, existing methods~\cite{embodiedqa,eqa_modular,yu2019multieqa} divide the entire process into two sub-tasks: an intermediate navigation task and a downstream question answering task. 

The navigation task is modeled as a partially observable Markov decision process.
The state space $\mathcal{S}$ consists of the positions and poses of the embodied agent during exploration which is not observable to the agent. The observation of the agent only contains the first-person-view images $\mathcal{O}$.
The state transition probability $P(s'|s, a)$ indicates the probability that an agent in the current state $s$ transfers to a new state $s'$ by an action $a$.
Let the initial state of an embodied agent be $s_0 = (x_0, y_0, \alpha_0$), where $x$ and $y$ indicate the spatial localization and $\alpha$ is the heading angle.
For the navigation task, the agent is expected to find a sequence of actions $\{a_0, a_1, ..., a_i \} \in \mathcal{A}$ based on the observations $\{o_0, o_1, ..., o_i\}$ and move closer to the target object $T$ specified in the question $q$.

The question answering task is performed after the exploration in the environment.
The agent is requested to answer the question $q$ based on its observation on the target $T$. 
The performance of the agent is evaluated by both the navigation performance (distance between the agent position and the target object) and the final question answering accuracy or QA accuracy for short.

\begin{figure}[t]
    \centering
    \includegraphics[width=\linewidth]{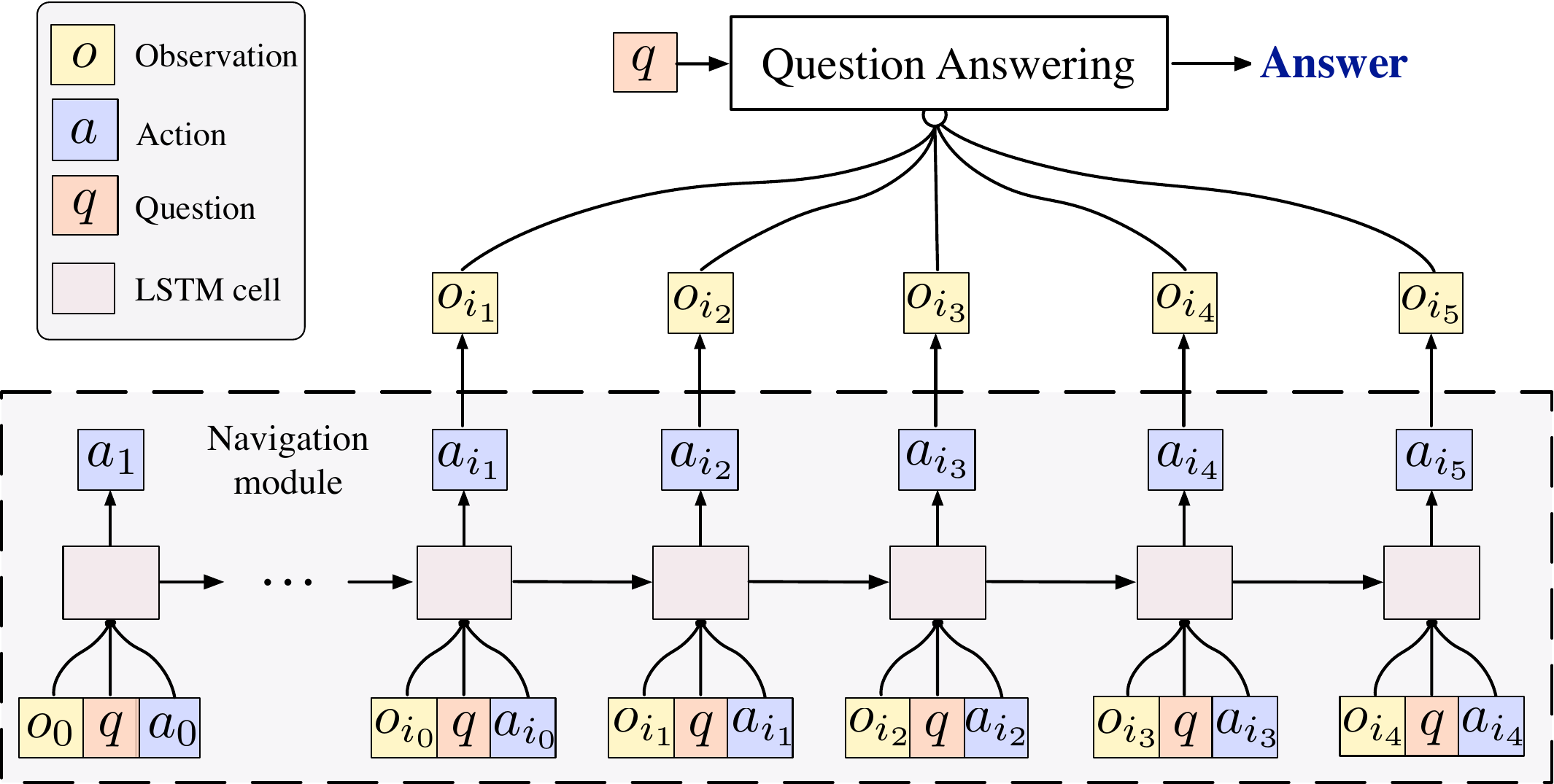}
    \caption{Overview of our proposed baseline. The model contains two modules, designed for the intermediate navigation task and downstream question answering task, respectively. We adopt the LSTM as the navigation model that takes as input the observation, current action, and question embedding.
    In training, the agent is placed on the shortest path from its spawning location to the target.
    At each step, the LSTM outputs a predicted action to be executed by the agent, which is supervised by mimicking the behavior of the shortest path. The QA module outputs the final answer based on the question and observations from the last five action steps.
    }
    \label{fig:model}
\end{figure}

\section{Baseline: A Simple pipeline without RL}
 \label{end2end}
 
For the ease of optimization, prior works~\cite{embodiedqa, eqa_modular} on EmbodiedQA treat the navigation and question answering as two separate modules and start training them separately. After that Deep RL is used to jointly train the navigation and the QA modules. In this section, we discuss a simple baseline network that can be jointly trained without RL. 

\textbf{Navigation module.} As opposed to directly minimizing the distance to the target via RL approaches, we regard the shortest path, from the spawn location to the target, as the supervision to guide the agent's behavior and train the navigation module using the imitation learning via SGD.

Formally, given a question and a current observation $o$, the navigation policy is represented by a model $\pi(a|q,o)$, where the policy $\pi$ generates a sequence of actions that move the agent.
We want to keep the model as simple as possible and use Long Short-Term Memory (LSTM)~\cite{hochreiter1997long} to model the policy and train the navigation module by mimicking the shortest path in a teacher-forcing way.
Given the observation input $o_i$, the question $q$, the current action $a_i$, and the current hidden state $h_{i}$, the policy $\pi$ predict the next action $a_{i+1}$ by updating the LSTM unit as follows:
\begin{align}
a_{i+1}, h_{i+1} = \texttt{LSTM} (o_i, q, a_i, h_{i}).
\end{align}
In training, the agent is placed on the shortest path and we expect the policy $\pi$ to output the same action as the shortest path does.
Therefore, the navigation problem can be optimized by the cross-entropy loss on the action prediction.
We found this simple navigation model performs better than other complex models that used before.

\textbf{Question answering module.}
Given the question $q$ and the sequence of observations from the last 5 steps along the navigation trajectory $\mathbf{o}=(o_{i-4}, o_{i-3}, ..., o_i)$,
the agent is expected to predict the answer to the input question,
\begin{align}
\texttt{ans}_i = \arg \max_{\texttt{ans}} p_\theta(\texttt{ans}|\mathbf{o}, q),
\label{eq:qa}
\end{align}
where $\texttt{ans}$ indicates the answer candidates and $p_\theta$ is the QA model.
Again we want to keep a simple model and reuse the same QA module in previous works~\cite{embodiedqa,eqa_modular}. The QA module encodes the question with a 2-layer LSTM and performs dot-product based attention between question embedding and image features from the last five frames along the navigation path right before the answer module is called.

\textbf{Joint training \textit{without} RL.} As shown in Fig.~\ref{fig:model}, we combine the navigation and the QA module together in a united SGD training pipeline. The concatenated embeddings of the observation, the current action, and the question are inputted into the navigation LSTM module. We initialize the model by mimicking the behavior of the shortest path navigation.
However, the inputs for the QA model are different for training and evaluation.
During training, the QA model is trained to maximize $p_\theta(\texttt{ans}|\mathbf{o}_{gt}, q)$, where $\mathbf{o}_{gt}$ is the observed images in the ground-truth path sequence. In evaluation, the QA model makes prediction using $p_\theta(\texttt{ans}|\mathbf{o}_{pred}, q)$, where $\mathbf{o}_{pred}$ is the observation at the predicted trajectory of the navigation model. 
To overcome the gap between training and evaluation, we draw the idea from Incremental Learning~\cite{ranzato2015sequence} and propose to gradually incorporate navigation predictions in training as the navigation model becomes more accurate. 

\begin{figure}[t]
    \centering
    \includegraphics[width=\linewidth]{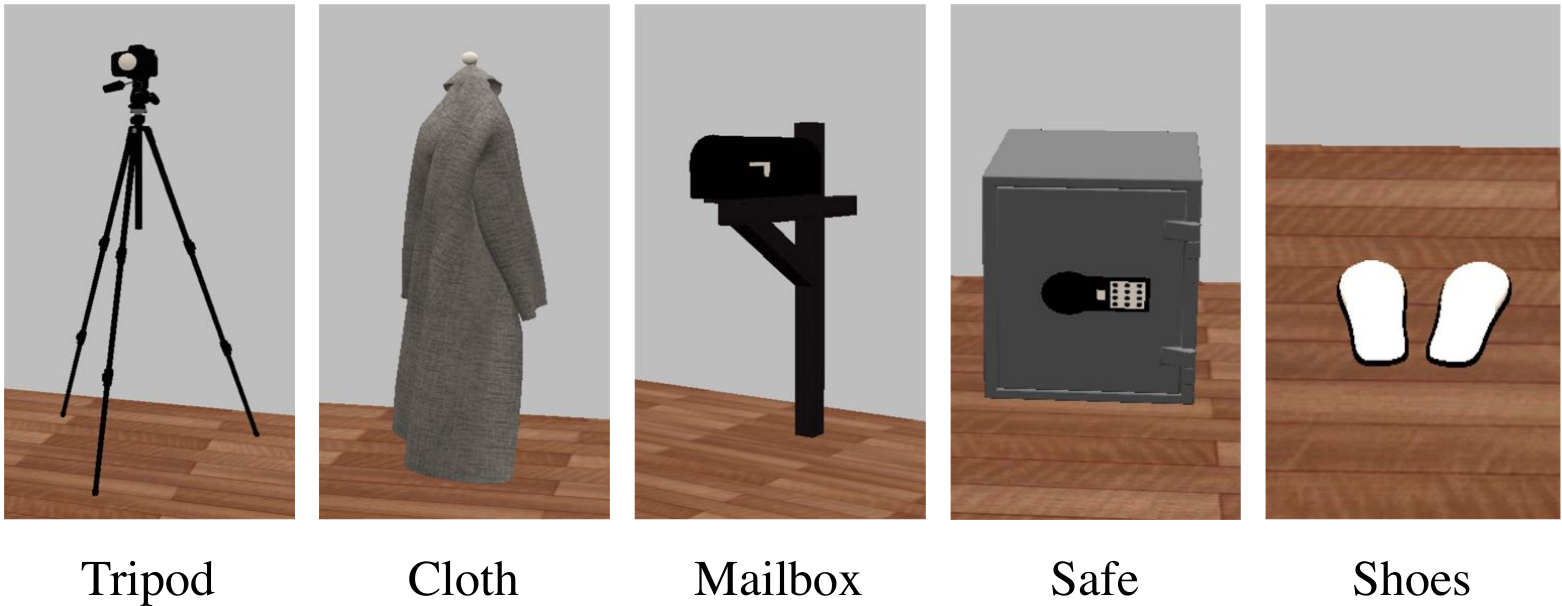}
    \caption{Examples of the five marker object candidates. We carefully select these objects from the SUNCG dataset~\cite{suncg} to avoid overlapping with the objects in the EQA v1 dataset~\cite{embodiedqa}.
    }
    \label{fig:markers}
\end{figure}

\section{Calibration Setting and Adaptation Method} \label{sec:marker}

In this section, we introduce an easier and practical and EmbodiedQA setting called \textit{calibration}.
We design a warm-up stage that the agent is asked a few rhetorical questions when it enters into a new environment. Our goal is to adapt the agent policy to the new environments. For example, when an agent is spawned in a new environment, a user places a few markers in a room and ask questions about the color of these markers. During the process, the agent learns to adapt its model parameters to the new environment.

We call it a \emph{calibration setting} as users could calibrate an agent to a new environment in a similar way as calibrating a new camera. The setting is essentially the same as the original setting except adding an extra QA step. It needs no additional supervision as the answers to the rhetorical questions are known. It is a practical setting as the adaption can be conveniently finished via the same QA interface without changing the agent's pipeline.

We propose a proxy task for the agent to explore the new environment by randomly placing some markers.
On top of that, we design a distillation framework for policy adaption.
We illustrate the way the agent explores environments with marker guidance, including how to set up markers and how to adapt the policy model to the new environments with these markers.

\subsection{Marker Generation}

We select the objects, which do not overlap with the objects in the training/validation/test objects in the EQA v1 dataset as our markers. In total, we found five such objects from the SUNCG dataset~\cite{suncg}, which is the source of the EQA v1 dataset.
The five markers are ``mailbox'', ``safe'', ``shoes'', ``tripod'', and ``cloth'', as shown in Fig.~\ref{fig:markers}. Note markers can be any foreign objects in practice and we use the above objects mainly because of their availability in the dataset. For simplicity, we do not use duplicate markers in the same environment.

\begin{figure}[t]
    \centering
    \includegraphics[width=\linewidth]{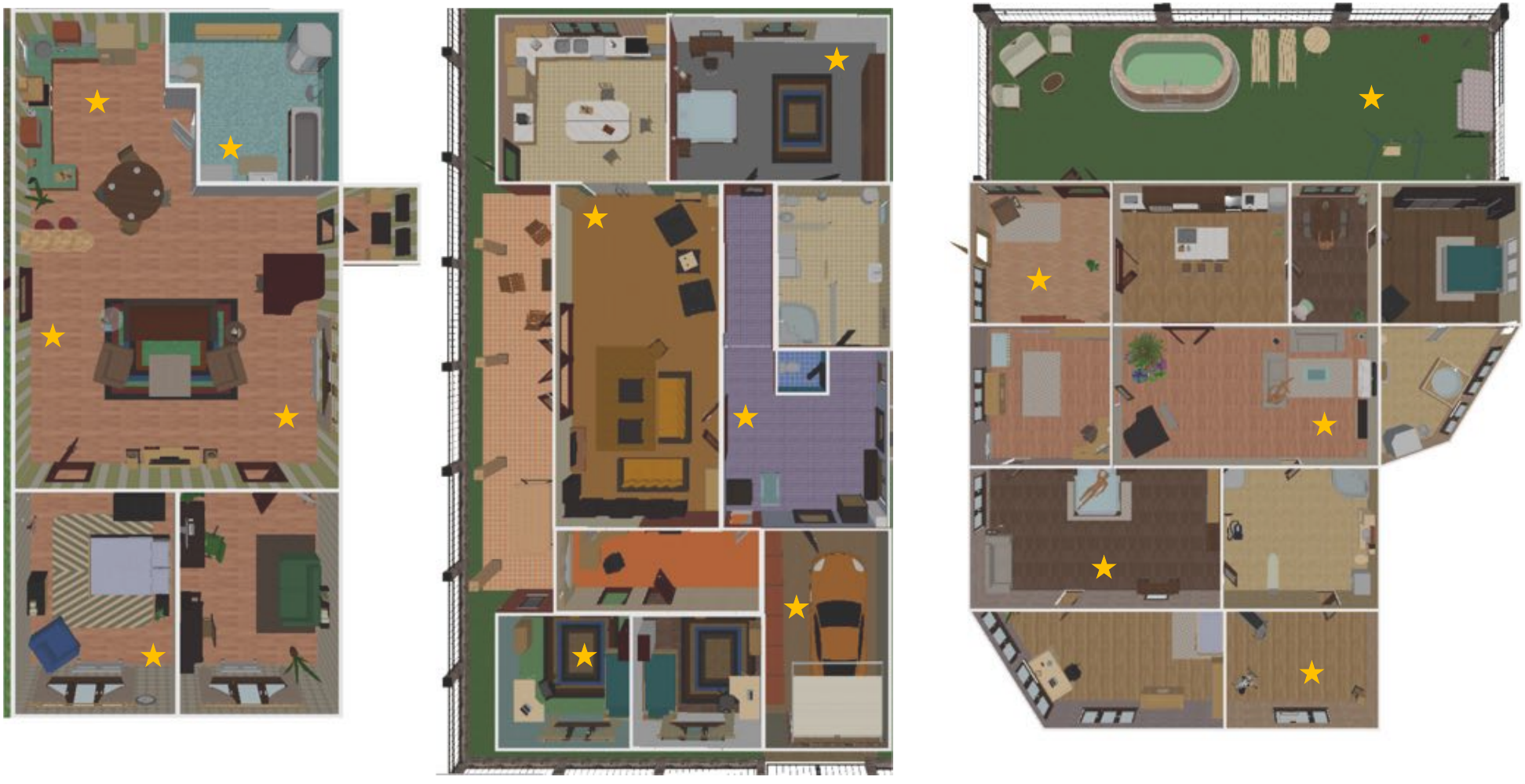}
    \caption{Examples of the placed positions of the five marker objects. Each yellow star indicates we place a marker there. 
    }
    \label{fig:markers_position}
\end{figure}

We randomly place the marker objects in each testing environment before evaluation. For example, each star in Fig.~\ref{fig:markers_position} indicates a location where we place a marker.
For each marker, we automatically generate a question: ``what is the color of the \texttt{<marker>}?''. 
We are able to paint arbitrary color on the placed marker objects so that answers to the questions do not matter.
We then generate the shortest path for a random spawn location to each target marker. In our experiments, the path is generated by the A$^\star$ search algorithm~\cite{astar} through the render API of the environment. Note as required in the EmbodiedQA setting, we are not allowed to access the structure data or global map about the environment during this process. In the case where A$^\star$ is unavailable, it is also feasible for users to mark the shortest path and input this information to the agent. In total, we generate 290 questions for 290 markers in 58 test environments. These questions will be used to adjust the agent during the adaptation stage. 

\subsection{Model Adaptation}

\begin{figure}[t]
    \centering
    \includegraphics[width=\linewidth]{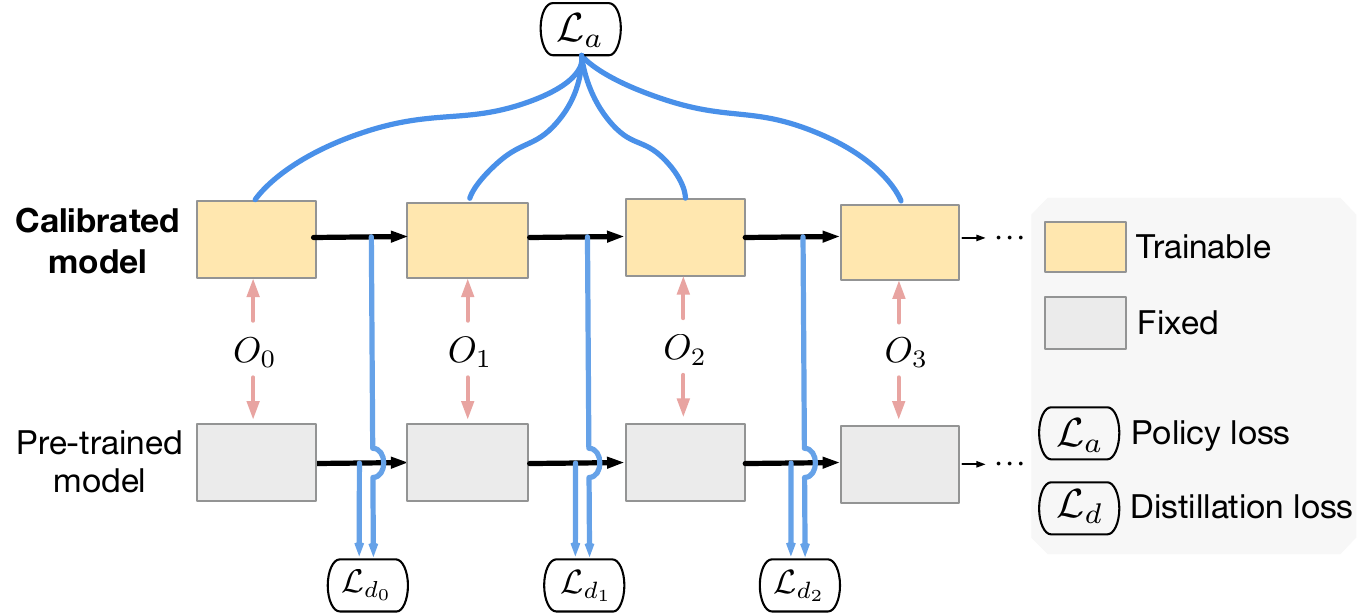}
    \caption{An illustration for the policy updating during calibration. The grey model in the figure is the trained navigation model from the training set, which is fixed during the adaptation process. The yellow model is the calibrated model that is jointly optimized by both the policy loss from imitation learning on marker data and the distillation loss on the intermediate representations from the pre-train model.
    }
    \label{fig:mqa_pipeline}
\end{figure}

We employ a policy distillation method for model adaptation.
Different from previous transfer learning methods~\cite{deng2019active},
our goal is to adapt the trained agent to a new test environment while retaining the essential knowledge learned from the original training set.
Inspired by recent work~\cite{luo2018graph}, we perform the distillation over the intermediate layer in the LSTM model.
Our adaptation network is shown in Fig.~\ref{fig:mqa_pipeline}. The input to this model is the automatically generated questions for the new environment. The model is learned to balance two types of supervision, \ie, the new policy loss of the markers and the imitation loss to the pre-trained representation. For convenience, we use the superscript to denote the adapted model. Similar to the previous training stage, the observation $o^c_i$, question embedding $q^c$ and current action $a^c_i$ are input to the navigation policy model at the $i$-th action step of the adaptation framework. We have the following modeling for the policy on the marker guided exploration,
\begin{align}
a^c_{i+1}, h^c_{i+1} = \texttt{Nav}^c (o^c_i, q^c, a^c_i, h^c_{i}).
\label{eq:lstm_c}
\end{align}

Let $\mathcal{L}^c_a$ be the policy loss over the new questions, defined by the cross-entropy loss on the action prediction with the action from the generated shortest path. 
Meanwhile during calibration, given the same input, we also minimize the distillation/imitation loss~\cite{hinton2015distilling} between the new (hidden states) and the pre-trained activations by:
\begin{align}
\mathcal{L}^c_d = \sum_i 1-\text{cos}(h^c_i, h^p_i),
\end{align}
where $\text{cos}$ denotes the cosine similarity in our experiments. $h^c_i$ and $h^p_i$ denote the hidden states of the adapted model and the pre-trained model at $i$-th step, respectively. The pre-trained model is held fixed during the distillation process.
The final adaptation loss function is computed from:
\begin{align}
\mathcal{L}^c = \lambda \mathcal{L}^c_d + (1-\lambda) \mathcal{L}^c_a,
\label{eq:final}
\end{align}
where the $\lambda$ is the parameter to balance the policy loss and distillation loss. We empirically analyze the value for $\lambda$ in Section~\ref{sec:ablation}. After the adaptation, the agent is tested on the test questions in the same manner as in the original setting.

\begin{table*}[!ht]
\setlength{\tabcolsep}{7pt}
\renewcommand{\arraystretch}{1.2}
\centering
\caption{Evaluation of navigation and question answering accuracy on the EQA v1 test set. * marks the agent uses no visual information.
``Test Envs'' indicates the test setting, where ``Standard'' indicates the agent's policy is fixed during evaluation. ``Calibration'' is the new setting where users can first ask the agent in the test environment a few questions about markers, during which the agent can change its policy.  We spawn the agent at 10, 30, and 50 action steps away from the target for evaluation, indicated $T_{-10}$, $T_{-30}$, and $T_{-50}$ in the table, respectively. $\mathbf{d_\Delta}$ measures the change, in the distance, towards the target after navigation. A higher $\mathbf{d_\Delta}$ indicates the agent is closer to the target when it stops.
}
\begin{tabular}{l | c | l c c c l| c c c l c c c l c c c }
\hline
\multirow{2}{*}{Methods}&\multirow{2}{*}{Test Envs}&&  \multicolumn{3}{c}{Navigation $\mathbf{d_\Delta}$} & & \multicolumn{3}{c}{QA $\mathbf{accuracy}$}\\
&&
 & \scriptsize$T_{-10}$ & \scriptsize$T_{-30}$ & \scriptsize$T_{-50}$ &
 & \scriptsize$T_{-10}$ & \scriptsize$T_{-30}$ & \scriptsize$T_{-50}$ \\
\hline
Blindfold*~\cite{anand2018blindfold} & Blind && -0.02* & -0.13* & -0.44* && 50.34\%* & {50.34}\%* & {50.34}\%* \\
\hline
PACMAN~\cite{embodiedqa} & Standard && -0.04  & 0.62 & 1.52 && 48.48\% & 40.59\% & 39.87\%\\
PACMAN +REINFORCE~\cite{embodiedqa} & Standard && 0.10 & 0.65 & 1.51 && 50.21\% & 42.26\% & 40.76\% \\
NMC~\cite{eqa_modular} & Standard && -0.29 & 0.73 & 1.21 && 43.14\% & 41.96\% & 38.74\% \\
NMC +A3C~\cite{eqa_modular} & Standard && 0.09 & 1.15 & 1.70 && \textbf{53.58}\% & 46.21\% & 44.32\% \\
\textbf{Ours} & Standard && \textbf{0.23} & \textbf{1.48} & \textbf{2.36} && 52.91\% & 48.01\% & 47.29\% \\
\hline
\textbf{Ours (Fine-tuning)} & Calibration && 0.13 & 1.52 & 3.27 && 52.11\% & 48.22\% & 48.27\% \\
\textbf{Ours (Distillation)}& Calibration && 0.19 & \textbf{1.53} & \textbf{3.39} && 52.23\% & \textbf{48.91\%} & \textbf{48.83\%} \\
\hline
\end{tabular}
\label{tab:state-of-the-art}
\end{table*}

\section{Experiments}
\subsection{Setup}

\noindent\textbf{Dataset and settings.} The EQA v1 dataset~\cite{embodiedqa} is a VQA dataset grounded in House3D~\cite{house3D}. 
There are 648 environments with 7,190 questions for training, 68 environments with 862 questions for validation and 58 environments with 933 questions for testing. The environments in different split have no overlapping. To sweep over the difficulty of the task, agents are set by spawning 10, 30, or 50 actions away from the target, denoted as $T_{-10}$, $T_{-30}$, and $T_{-50}$, which corresponds to the distance of 0.94, 5.39, and 11.01 meters from the target, respectively. For the medium and long distance ($T_{-30}$, $T_{-50}$) questions, the target objects are usually not visible to the agent, and hence the agent needs to take actions to explore the environment to reach the target.

In the \emph{standard} setting, the agent is optimized only on the training data split. Once trained, its policy is fixed and then evaluated on the test questions.
In our proposed \emph{calibration} setting, users can ask the agent a few questions about markers in the test environment. During the process, the agent can change its policy, and its policy is fixed thereafter and evaluated on the test questions.

\noindent\textbf{Baselines.}
We compare our method to the following baselines: (i) \emph{Planner-Controller Navigation Module (PACMAN)}~\cite{embodiedqa} designs a hierarchical structure for navigation module, with a planner to select actions (directions) and a controller that decide how far to move following this action. The navigation module and visual question answering model are first trained separately and then jointly trained by the reinforcement learning method. (ii) \emph{Neural Modular Control (NMC)}~\cite{eqa_modular} proposes a hierarchical policy for the navigation module, where the higher-level master policy proposes subgoals to be executed by sub-policies. (iii) \emph{Blindfold baseline}~\cite{anand2018blindfold} is a question-only BoW baseline that never navigates nor observes the environment.
We cite their performance reported in their papers for an apple-to-apple comparison.

\begin{figure*}[t]
    \centering
    \includegraphics[width=\linewidth]{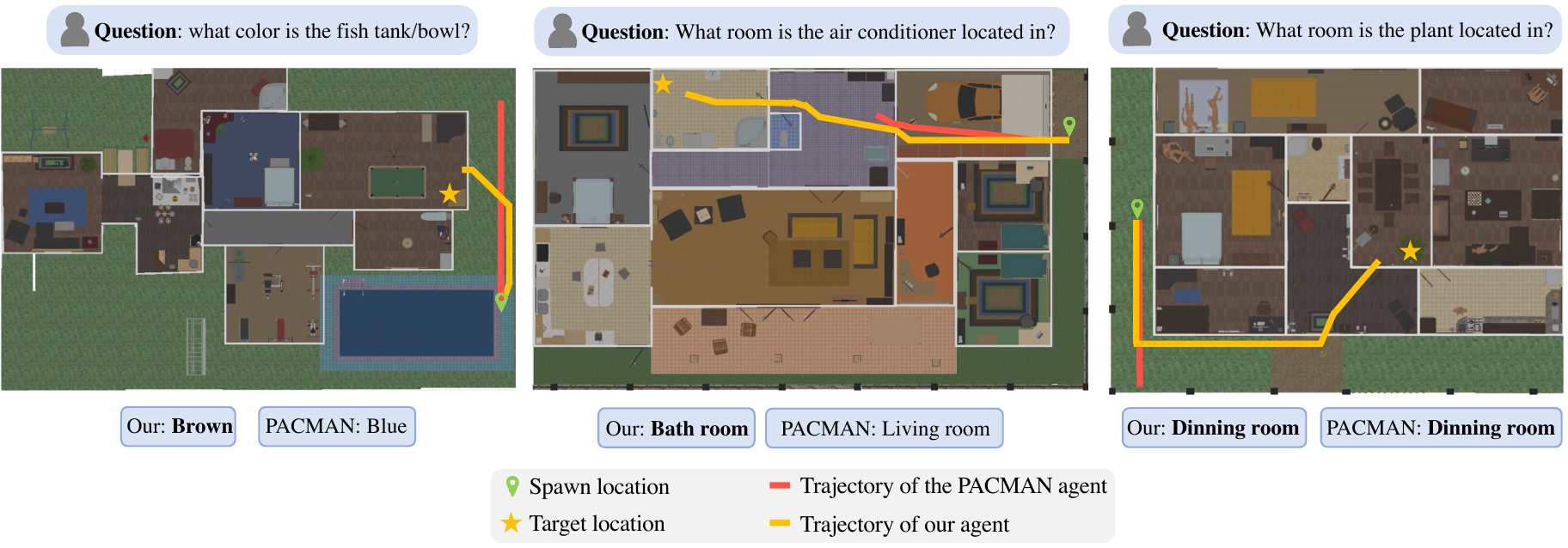}
    \caption{Comparison of qualitative results with PACMAN~\cite{embodiedqa} on the EQA v1 dataset. 
    The yellow line and red line indicate the trajectories of our calibrated agent and PACMAN agent, respectively. 
    As we can see, given a question about a specific object (indicated by the star in the figure), 
    our calibrated agent could easily find the correct direction and get close to the target for further observing. In contrast, the previous PACMAN agent has no prior information about the house, resulting in the wrong direction to explore.
    The answers of the two agents are shown at the bottom of each environment. The answer in bold is the correct answer to the question.
    }
    \label{fig:visual}
\end{figure*}

\noindent\textbf{Evaluation Metrics.}
Following~\cite{embodiedqa,eqa_modular}, we evaluate the performance of the navigation and the question answering accuracy. We evaluate the navigation policy by \dD, which is the change in the distance towards the target, from the initial spawning position to the final stopping position. In other words, how much progress does the agent make moving towards the target.
As mentioned, the EmbodiedQA agent is expected to first observe the target before answering the question about it. 
The navigation performance \dD~is, therefore perhaps, equally important in evaluating an agent. There are 191 answer candidates in the EQA v1 dataset. Similar to other VQA tasks, the question answering performance is measured by the classification accuracy.

\noindent\textbf{Implementation details.} We use PyTorch~\cite{paszke2017pytorch} for all experiments. We develop our model based on the code released by~\cite{embodiedqa}. In experiments, we take the pre-trained CNN model released by \cite{embodiedqa} as the image encoder.
The input question is embedded by a two-layer LSTM with cell size 64, and we learn the embedding for each word with 64 dimensions.
We take a two-layer LSTM with cell size 128 as the navigation module.
The $\lambda$ in Eqn.~\eqref{eq:final} is set to be 0.2 according to the performance on the validation set.
The Adam~\cite{kingma2014adam} optimizer with the learning rate of $1 \times 10^{-5}$ is used for all the experiments. The source code and models will be made available to the public. 

\subsection{Baseline Comparison}
Table~\ref{tab:state-of-the-art} summarizes the comparison in terms of the navigation and the QA accuracy on the EQA v1 dataset. Recall Blindfold is the BOW baseline in~\cite{anand2018blindfold} using no visual information. Even though the Blindfold baseline achieves the best QA accuracy among all of the other methods, it suffers from the very poor (random) navigation accuracy. This is expected since the agent is merely learning the language bias between the question and answers. In contrast, other EmbodiedQA methods achieve much higher navigation accuracy which lends credibility to their actual QA performance. In the standard setting, our simple baseline method achieves comparable or even better performance compared to state-of-the-art methods such as PACMAN and NMC. It yields the best navigation accuracy and competitive QA performance. The improvement on QA accuracy is more evident for the medium ($T_{-30}$) and long ($T_{-50}$) distance questions. These results verify the efficacy of our LSTM navigation policy over other more complicated policies.

Under the calibration setting, ``ours (distillation)'' method outperforms our baseline method. The biggest gain comes from the navigation, especially from the medium distance ($T_{-30}$) and long distance ($T_{-50}$) navigation. For example, with the calibration, the long distance ($T_{-50}$) evaluation is improved from 2.36 to 3.39, and the QA accuracy is also increased from 47.29\% to 48.83\%. The results substantiate the importance of the introduced calibration step in the EmbodiedQA setting. The comparison between the last two rows suggests the distillation may be more effective than fine-tuning in this calibration. It is worthwhile noting that the calibration seems not improving the performance for the short distance questions ($T_{-10}$). In this situation, the agent is spawned very closed to the questioned object (on average 0.94 meters), and the agent is often able to observe the target without executing any further move.

In both settings, we should note that even though our simple baseline achieves competitive results, EmbodiedQA remains to be a very challenging task. The room for further improvement is still very big. We believe future researches will significantly beat this simple baseline both in terms of navigation and QA accuracy.

\begin{figure}[t]
    \centering
    \includegraphics[width=\linewidth]{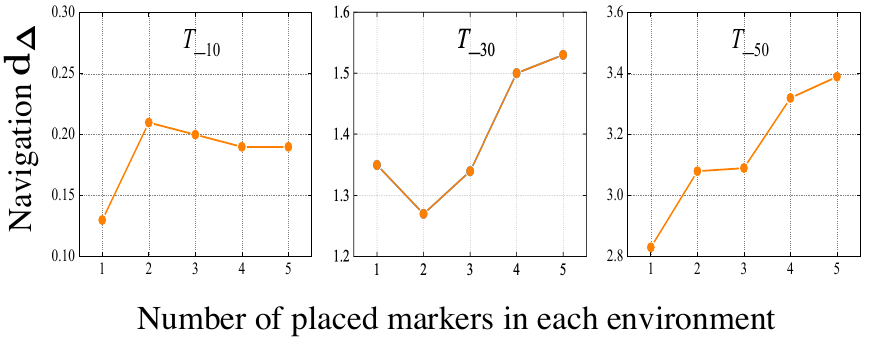}
    \caption{Navigation performance curve over the number of placed markers in each environment. We evaluate the agent at the short distance $T_{-10}$, medium-distance $T_{-30}$, and long-distance $T_{-50}$. The curves prove the effectiveness of the placed markers. In general, the more markers we place, the higher the navigation performance is achieved.}
    \label{fig:num_markers}
\end{figure}

\subsection{Qualitative Results}
We show some qualitative results in Fig.~\ref{fig:visual}. The yellow line and red line indicate the trajectories of our agent and PACMAN~\cite{embodiedqa} agent, respectively. As we can see, given a question about a specific object (indicated by the star in the figure), 
our agent can find the correct direction to explore and usually moves closer to the target. In contrast, the previous PACMAN agent has no idea about where the target is in this new environment. As a result, the agent may behave randomly, wasting the valuable action steps, or explore in a totally wrong direction. 

The answer predictions are shown at the bottom of each environment. The answer in bold indicates the correct answer. For the rightmost question, although PACMAN predicts the correct answer, the agent is actually quite far away from the target, \ie~its prediction is right but not for the right reason. The result echoes the statistical irregularities in the EQA v1 dataset as reported in~\cite{anand2018blindfold}. However, we believe this issue can be mitigated by evaluating both QA and navigation accuracy.

\subsection{Setting Studies} \label{sec:ablation}
We design experiments to evaluate our design choices in the proposed calibration setting.

\noindent\textbf{The number of placed markers.}
As introduced in Section~\ref{sec:marker}, we select five markers from the SUNCG dataset that do not overlap with the target objects in the EQA v1 dataset. By default, we place all of them in each test environment.
To understand the influence of the placed markers on the navigation performance, we gradually place the markers in each environment.
Fig.~\ref{fig:num_markers} shows the navigation performance curves over the number of placed markers in the $x$-axis. 
These results indicate placing more markers improve the agent's performance for medium-($T_{-30}$) and long-distance ($T_{-50}$) questions. The results suggest that users can repeatably place markers more time to improve the agent's performance. It is less beneficial for the short-distance questions where the agent is often able to see the target at its spawn location.

\begin{figure}[t]
    \centering
    \includegraphics[width=0.98\linewidth]{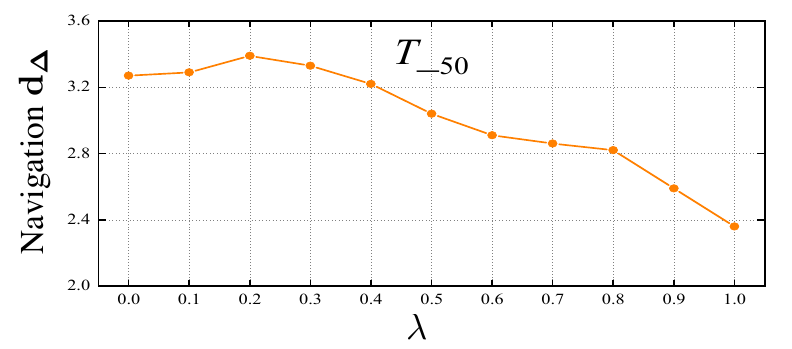}
    \vspace{-2mm}
    \caption{Performances on different values of the balancing parameter $\lambda$ defined in Eqn.~\eqref{eq:final}.} 
    \vspace{-3mm}
    \label{fig:lambda}
\end{figure}

\noindent\textbf{The balancing parameter $\mathbf{\lambda}$.}
In Eqn.~\eqref{eq:final}, $\lambda$ is a hyper-parameter balancing the policy learned from the training data and the new environment.
Fig.~\ref{fig:lambda} shows the performance curves over different values of $\lambda$, in which $\lambda=0$ denotes only using the policy loss whereas $\lambda=1$ is only distilling from the pre-trained model. From the difference between $\lambda=0$ and $\lambda=1$, we see that the calibration help improve the pre-trained model. The best navigation performance is achieved at $\lambda=0.2$ which suggests our distillation can balance the two losses and perform better than using the either loss alone.

\section{Conclusion}
In this paper, we conducted an empirical study to revisit a recent topic of EmbodiedQA. Our studies show that EmbodiedQA is an interesting yet challenging problem where the QA and navigation accuracy are both needed to faithfully evaluate an agent. 
Besides, we show that a simple yet effective joint training baseline is able to achieve very competitive results to the state-of-the-art.
We found the poor navigation in a new environment is the current bottleneck for EmbodiedQA. We tackled this issue from another angle by introducing a slightly different ``calibration'' setting. In this new setting, to assist the agent in adapting to the new environment, we design a proxy task where we randomly place markers and ask questions about them. These rhetorical questions drive the agent to explore and familiarize the environment. Experimental results validate this small change in setting leads to a significant gain in navigation when the agent is further (30 and 50 action steps) away from the target.

{
\bibliographystyle{IEEEtran}
\bibliography{egbib}
}

\begin{IEEEbiography}[{\includegraphics[width=1in,height=1.25in,clip,keepaspectratio]{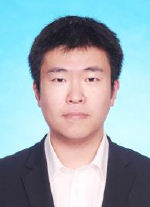}}]{Yu Wu} received the B.E. degree from Shanghai Jiao Tong University, China, in 2015. He is currently a Ph.D. candidate in the Center for Artificial Intelligence, University of Technology Sydney, Australia. His research interests are video analysis and person re-identification.    
\end{IEEEbiography}

\begin{IEEEbiography}[{\includegraphics[width=1in,height=1.25in,clip,keepaspectratio]{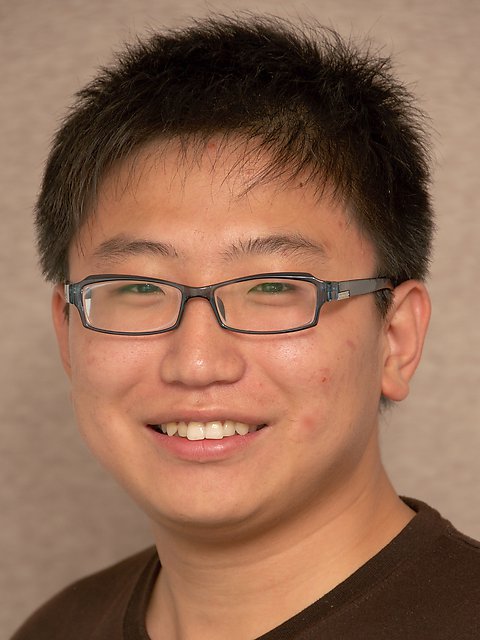}}]{Lu Jiang} is a senior research scientist at Machine Perception in Google Research. He received the Ph.D. in Artificial Intelligence (Language Technology) from Carnegie Mellon University in 2017. Lu's primary interests lie in the interdisciplinary field of Multimedia, Machine Learning, and Computer Vision. He is the recipient of Yahoo Fellow and Erasmus Mundus Scholar. He received Best paper nomination at ACL and ACM ICMR; Best poster at IEEE SLT; Best system in NIST TRECVID.
\end{IEEEbiography}
    
\begin{IEEEbiography}[{\includegraphics[width=1in,height=1.25in,clip,keepaspectratio]{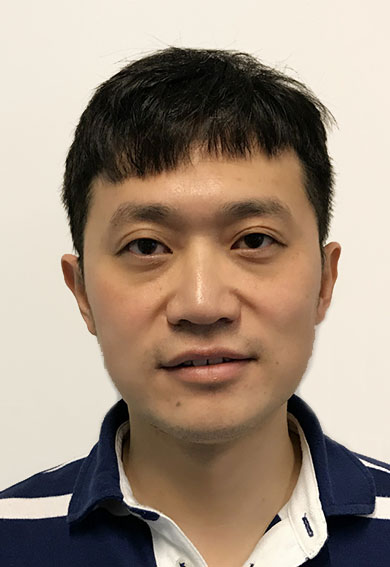}}]{Yi Yang} received the Ph.D. degree in computer science from Zhejiang University, Hangzhou, China, in 2010. He is currently a professor with the University of Technology Sydney, Australia. He was a post-doctoral researcher in the School of Computer Science, Carnegie Mellon University, Pittsburgh, Pennsylvania. His current research interests include machine learning and its applications to multimedia content analysis and computer vision, such as multimedia indexing and retrieval, surveillance video analysis, and video content understanding.    
\end{IEEEbiography}
\end{document}